\documentclass[10pt]{article}

\usepackage[utf8]{inputenc}
\usepackage{graphicx}
\usepackage{amsmath, amssymb}
\usepackage{booktabs}
\usepackage{multirow}
\usepackage{subcaption}
\usepackage{hyperref}
\usepackage{siunitx}
\usepackage{float}
\usepackage{xcolor}
\usepackage{url}
\usepackage{microtype}
\usepackage{tikz}
\usetikzlibrary{positioning, arrows.meta}
\usepackage{pgfplots}
\pgfplotsset{compat=1.18}

\hypersetup{
    colorlinks=true,
    linkcolor=blue,
    citecolor=blue,
    urlcolor=blue
}

\title{Weakly Supervised Continuous Micro-Expression Intensity Estimation \\
Using Temporal Deep Neural Networks}

\author{
    Riyadh Almushrafy \\
    \small Department of Computer and Information Sciences, \\
    \small College of Science at Zulfi, Majmaah University, \\
    \small Al-Majmaah 11952, Saudi Arabia \\
    \small \texttt{r.almushrafy@mu.edu.sa}
}

\date{}

\begin{document}

\maketitle

\begin{abstract}
Micro-facial expressions (MFEs) are brief, subtle, and involuntary facial movements that reveal genuine affective states. While most prior work has focused on recognizing discrete micro-expression categories, relatively few studies have examined the \emph{continuous} evolution of micro-expression intensity. Progress in this direction has been limited by the absence of frame-level intensity annotations in existing datasets, which makes fully supervised regression impractical.

This paper presents a unified, dataset-agnostic framework for continuous micro-expression intensity estimation using only weak temporal labels (onset, apex, offset). A triangular prior is used to convert sparse temporal landmarks into dense pseudo-intensity trajectories, and a lightweight temporal regression model—combining a ResNet18 encoder with a bidirectional GRU—is trained to predict frame-wise intensity directly from raw images. The method requires no frame-level annotation effort and is applied consistently across datasets through a common preprocessing and temporal alignment pipeline.

Experiments on SAMM and CASME~II show that the proposed model achieves strong temporal agreement with the pseudo-intensity trajectories. On SAMM, the ResNet18–GRU model reaches a Spearman correlation of \textbf{0.9014} and Kendall’s~$\tau$ of \textbf{0.7999}, outperforming a frame-wise baseline. On CASME~II, the model achieves up to \textbf{0.9116} in Spearman’s~$\rho$ and \textbf{0.8168} in Kendall’s~$\tau$ when trained without the apex-ranking term, which aligns with the dataset’s weaker apex dominance. Ablation studies indicate that temporal modeling and structured pseudo labels play a central role in capturing the rise–apex–fall dynamics of MFEs.

Based on available literature, this appears to be the first unified approach for continuous micro-expression intensity estimation using only sparse temporal annotations, offering a practical foundation for fine-grained affective behavior analysis under limited supervision.
\end{abstract}

\section{Introduction}

Micro-expressions are brief, involuntary facial movements that reveal genuine
emotional states even when individuals attempt to suppress them. Their very
short duration (typically 1/25--1/5\,s) and low intensity make them difficult to
analyze, particularly in real-world settings where high-speed imaging and dense
annotations may not be available. Although most existing work focuses on
categorical micro-expression recognition, estimating how expression intensity
changes over time provides a more detailed and potentially more informative
representation of facial behaviour.

A major challenge for continuous intensity estimation is the absence of
frame-level intensity annotations in current micro-expression datasets. Available
datasets typically provide only three temporal landmarks---onset, apex, and
offset---which convey the overall structure of the expression but not its
frame-wise intensity profile. This limitation motivates weakly supervised
approaches that make use of sparse temporal cues to approximate the underlying
intensity trajectory.

This paper presents a unified framework for continuous micro-expression
intensity estimation under weak supervision. The approach uses a triangular
pseudo-intensity trajectory derived from onset--apex--offset annotations and
learns to regress this trajectory through a combination of spatial encoding and
bidirectional temporal modeling. A lightweight ResNet18 backbone extracts
per-frame features, and a bidirectional GRU models the rapid rise and decay
characteristic of micro-expressions.

Experiments indicate that temporal recurrence accounts for most of the model’s
performance: the GRU alone captures the essential temporal structure, achieving
correlations above 0.97 on SAMM and 0.98 on CASME~II. Auxiliary losses such as
smoothness and apex-ranking act mainly as mild regularizers and have limited
impact on final results. These observations suggest that the triangular
pseudo-label provides a stable supervisory signal and that temporal modeling is
well suited to the weak-supervision setting.

\textbf{Contributions.} This work provides:

\begin{itemize}
    \item A weakly supervised framework for continuous micro-expression
    intensity estimation using triangular pseudo-labels derived from sparse
    temporal annotations.

    \item A temporally aware architecture that combines a lightweight spatial
    encoder with a bidirectional GRU to model fine-grained temporal dynamics.

    \item A set of experiments and ablations showing that temporal recurrence is
    the main contributor to performance, with auxiliary losses serving as
    optional regularizers.

    \item A reproducible pipeline that operates consistently across datasets
    without requiring dataset-specific adjustments.
\end{itemize}

Together, these components form a practical approach for continuous
micro-expression intensity estimation under realistic annotation constraints.

\section{Related Work}

Research on micro-expression analysis spans several related areas, including
dataset development, categorical micro-expression recognition, facial action unit
(AU) intensity estimation, and weakly supervised learning. This section reviews
the most relevant work and positions the present study within this context.

\subsection{Micro-Expression Datasets}

Early spontaneous micro-expression datasets such as
SMIC~\cite{smic_dataset} and CASME~\cite{casme} established initial benchmarks
but provided limited spatial resolution and relatively coarse temporal
annotations. Subsequent datasets—most prominently CASME~II~\cite{casme2} and
SAMM~\cite{samm}—introduced high-frame-rate recordings (200\,fps) together with
precise onset, apex, and offset labels, enabling more detailed temporal
analysis. However, none of these datasets include frame-level intensity
annotations, which limits the use of direct supervised regression and motivates
weakly supervised alternatives.

\subsection{Micro-Expression Recognition}

Most prior work addresses categorical micro-expression recognition. Earlier
methods relied on handcrafted spatio-temporal descriptors such as
LBP-TOP~\cite{lbp_top}. More recent deep learning approaches include
optical-flow CNNs~\cite{li2019microexpression}, dual-stream attention
architectures~\cite{liu2021dual}, and multi-scale temporal convolutional
networks~\cite{peng2019dual}. These methods capture subtle motion patterns but
operate at the sequence level and do not produce continuous frame-wise intensity
trajectories. They also typically assume pre-segmented clips and do not
explicitly model the characteristic rise–apex–fall structure of
micro-expressions.

\subsection{Facial Action Unit Intensity Estimation}

Continuous expression intensity estimation has been studied more extensively in
the AU domain, supported by datasets such as DISFA~\cite{disfa} and
BP4D~\cite{bp4d}. Existing methods include patch-based regression
approaches~\cite{zhao2016joint}, deep multi-task models~\cite{kollias2021deep},
and graph-based spatio-temporal networks~\cite{zhang2022graph}. Although these
studies provide useful insights, AU datasets primarily contain longer and
higher-amplitude macro-expressions with dynamics that differ from those of
micro-expressions. Moreover, dense AU labels are costly to obtain and are not
well suited to high-frame-rate micro-expression sequences.

\subsection{Pseudo-Labeling and Weak Supervision}

Pseudo-labeling has been explored for micro-expression intensity estimation.
Zhang et al.~\cite{zhang2023micro} proposed sparse temporal pseudo-labels
centered on the apex frame, though these cues are limited in temporal extent and
depend on dataset properties. Lin et al.~\cite{lin2023temporal} employed
temporal interpolation to approximate frame-wise intensity, but the resulting
trajectories are not explicitly motivated by expression dynamics and may vary
across datasets.

Pseudo-labeling also appears widely in weakly supervised vision. Lee~\cite{lee2013pseudolabel}
showed that dense labels can be inferred from sparse anchors, and Wei
et al.~\cite{wei2018weak} demonstrated iterative refinement of weak cues for
object detection and segmentation. In AU intensity estimation, inferred or
partial labels have been used as surrogate supervision~\cite{zhao2016joint,
kollias2021deep}. These works indicate that temporal structure can be learned
without access to dense annotations.

However, existing micro-expression approaches rely on sparse local cues,
interpolation-based heuristics, or dataset-specific assumptions, and do not
provide a general pseudo-intensity trajectory that can be applied consistently
across datasets.

\subsection{Summary}

This work builds on prior research by introducing a dataset-agnostic
pseudo-intensity formulation based solely on onset--apex--offset annotations.
The triangular trajectory offers a simple structured signal for supervision, and
when combined with temporal modeling, enables frame-wise intensity estimation
under weak annotation conditions.

\section{Methodology}
\label{sec:method}

This section describes the proposed framework for continuous micro-expression
intensity estimation under weak supervision. The objective is to predict a dense
frame-level intensity trajectory using only onset, apex, and offset landmarks.
The framework consists of four components:
(i) a dataset-independent annotation and preprocessing stage,
(ii) a triangular pseudo-intensity formulation,
(iii) a convolutional–recurrent regression model, and
(iv) a loss function designed for weakly supervised learning.

\subsection{Unified Annotation and Preprocessing Pipeline}

Micro-expression datasets differ in directory structure, naming conventions, and
annotation formats. To ensure consistency across datasets, all clips are
converted into a common annotation format containing the subject ID, clip ID,
ordered frame paths, and temporal landmarks (onset, apex, offset). Optional
metadata such as action units (AUs) or emotion labels may be included but are
not used during training.

\begin{figure}[!t]
\centering
\begin{tikzpicture}[>=latex, node distance=1.4cm, thick]

\tikzstyle{frame} = [rectangle, draw, rounded corners,
    minimum width=0.9cm, minimum height=0.8cm, align=center]
\tikzstyle{block} = [rectangle, draw, rounded corners,
    minimum width=2.6cm, minimum height=1.0cm, align=center]
\tikzstyle{arrow} = [->, thick]

\node[frame] (x1) {$x_1$};
\node[frame, right=0.2cm of x1] (x2) {$x_2$};
\node[frame, right=0.2cm of x2] (x3) {$\cdots$};
\node[frame, right=0.2cm of x3] (xtm1) {$x_{T-1}$};
\node[frame, right=0.2cm of xtm1] (xT) {$x_T$};

\node[above=0.05cm of x3] {\small $T$ resampled frames (onset $\rightarrow$ offset)};

\node[frame, below=1.3cm of x2] (xon) {\includegraphics[width=0.9cm]{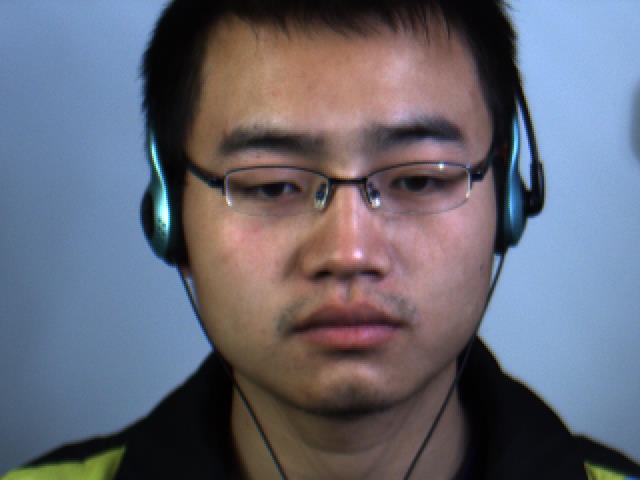}};
\node[frame, below=1.3cm of x3] (xap) {\includegraphics[width=0.9cm]{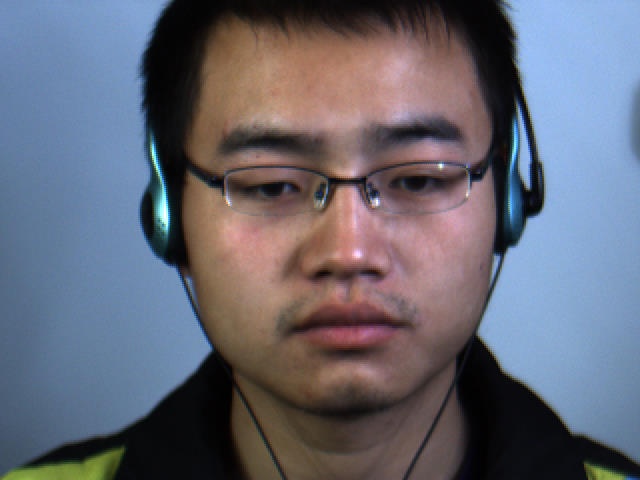}};
\node[frame, below=1.3cm of xtm1] (xoff) {\includegraphics[width=0.9cm]{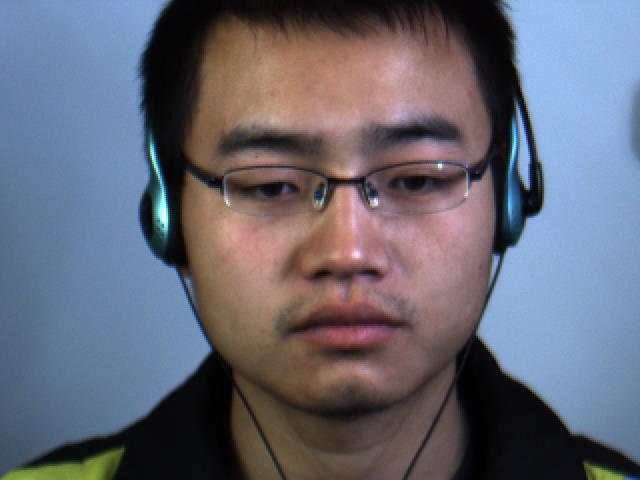}};

\node[below=0.25cm of xon] {\scriptsize Onset};
\node[below=0.25cm of xap] {\scriptsize Apex};
\node[below=0.25cm of xoff] {\scriptsize Offset};

\node[above left=0.0cm and -1.1cm of xap] {\small Example subject frames};

\draw[dashed] (x1.south) -- ++(0,-0.35) -- (xon.north);
\draw[dashed] (x3.south) -- ++(0,-0.35) -- (xap.north);
\draw[dashed] (xtm1.south) -- ++(0,-0.35) -- (xoff.north);

\node[block, below=2.0cm of xap] (resnet) {ResNet18\\(spatial encoder)};

\draw[arrow] (xon.south) -- ++(0,-0.25) -| (resnet.north);
\draw[arrow] (xap.south) -- ++(0,-0.25) -- (resnet.north);
\draw[arrow] (xoff.south) -- ++(0,-0.25) -| (resnet.north);

\node[block, below=1.8cm of resnet] (gru) {Bidirectional GRU\\(temporal encoder)};
\draw[arrow] (resnet.south) -- (gru.north);

\node[block, below=1.7cm of gru] (reg) {Linear layer\\(frame-wise regression)};
\draw[arrow] (gru.south) -- (reg.north);

\node[block, below=1.7cm of reg, minimum width=4.8cm] (traj)
    {Predicted trajectory $\hat{y}(t)$};
\draw[arrow] (reg.south) -- (traj.north);

\node[block, right=3.6cm of gru] (triangle)
    {Triangular pseudo\\trajectory $y(t)$};

\node[block, right=3.6cm of resnet] (landmarks)
    {Onset–Apex–Offset\\annotations};

\draw[arrow] (landmarks.south) -- (triangle.north);
\draw[arrow] (triangle.west) -- node[midway, above]
    {\small supervision} (traj.east);

\end{tikzpicture}

\caption{Overview of the proposed weakly supervised framework. 
\emph{Top:} a micro-expression clip is represented by $T$ uniformly resampled 
frames between onset and offset. \emph{Middle:} example onset, apex, and offset 
frames taken from publication-permitted CASME~II subjects (© Xiaolan Fu). 
\emph{Bottom:} all $T$ frames are encoded by a ResNet18 backbone, aggregated 
temporally by a bidirectional GRU, and mapped to a continuous intensity 
trajectory supervised by triangular pseudo-labels derived from 
onset--apex--offset landmarks.}

\label{fig:framework}
\end{figure}
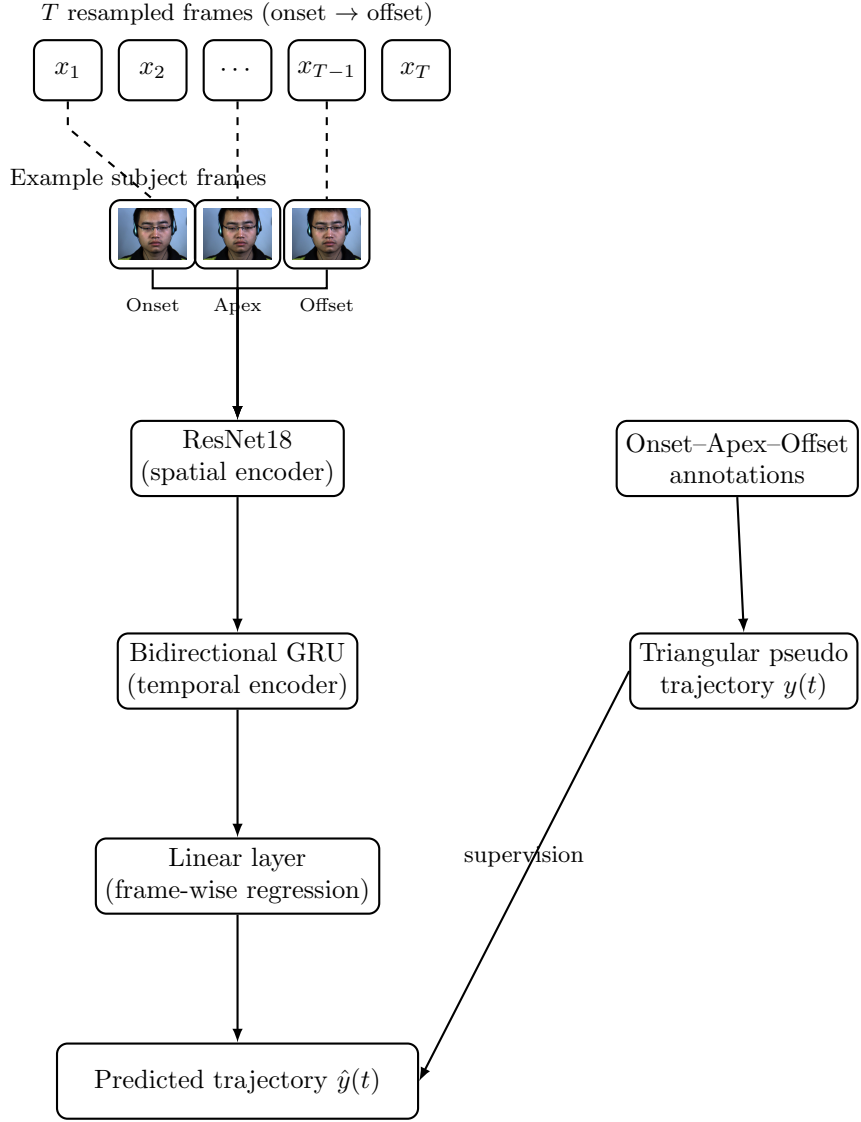

An overview of the framework is shown in Figure~\ref{fig:framework}. Because
micro-expression durations vary across subjects and datasets, each sequence is
uniformly resampled to $T$ frames between the annotated onset and offset
positions. This temporal normalization preserves relative timing while providing
a fixed-length input for the temporal model.

\subsection{Triangular Pseudo-Intensity Trajectories}
\label{sec:triangle}

Because existing micro-expression datasets do not provide frame-level intensity
annotations, we construct a \emph{triangular pseudo-intensity trajectory} to
approximate the rise–apex–fall behaviour commonly observed in micro-expressions.
This provides dense supervision from sparse temporal landmarks.

\begin{figure}[!t]
\centering
\begin{tikzpicture}
\begin{axis}[
    width=0.8\linewidth,
    height=5cm,
    xmin=0, xmax=1,
    ymin=0, ymax=1.05,
    xlabel={Normalized time $\tau(t)$},
    ylabel={Pseudo intensity $y(t)$},
    xtick={0,0.3,0.7,1},
    xticklabels={$0$, $\tau_{\mathrm{on}}$, $\tau_{\mathrm{ap}}$, $\tau_{\mathrm{off}}$},
    ytick={0,1},
    yticklabels={$0$, $1$},
    axis lines=left,
    domain=0:1,
    samples=200,
    thick
]
\addplot[very thick] coordinates {
    (0,0)
    (0.7,1)
    (1,0)
};
\end{axis}
\end{tikzpicture}

\caption{Triangular pseudo-intensity trajectory derived from onset–apex–offset
landmarks. This prior provides dense supervision despite the absence of
frame-level intensity labels.}
\label{fig:triangle}
\end{figure}
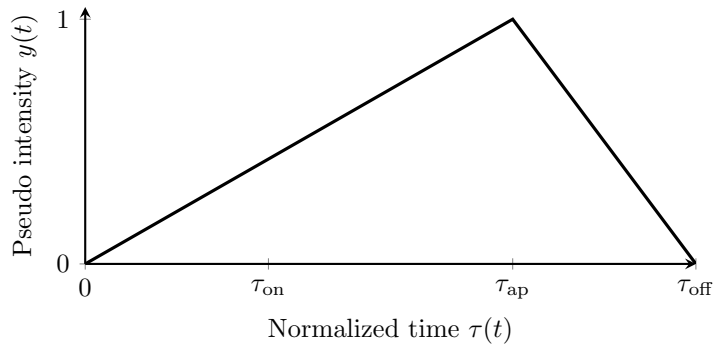

Figure~\ref{fig:triangle} illustrates the triangular structure. Let
$(f_{\mathrm{on}}, f_{\mathrm{ap}}, f_{\mathrm{off}})$ denote the annotated onset,
apex, and offset frames. For uniformly sampled frames indexed by
$t \in \{0, \dots, T{-}1\}$, the normalized time is defined as:
\begin{equation}
    \tau(t) = \frac{t}{T - 1}.
\end{equation}

The apex location in normalized time is:
\begin{equation}
    \alpha = \frac{f_{\mathrm{ap}} - f_{\mathrm{on}}}{f_{\mathrm{off}} - f_{\mathrm{on}}}.
\end{equation}

The pseudo-intensity target is then:
\begin{equation}
    y(t) =
    \begin{cases}
        \frac{\tau(t)}{\alpha + \varepsilon}, & \tau(t) \le \alpha, \\
        \frac{1 - \tau(t)}{1 - \alpha + \varepsilon}, & \tau(t) > \alpha,
    \end{cases}
    \label{eq:triangular}
\end{equation}
yielding $y(t) \in [0,1]$. This formulation provides a simple, interpretable, and
consistent supervisory signal across datasets.

\subsection{Temporal Intensity Regression Model}

As illustrated in Figure~\ref{fig:framework}, the regression model combines a
spatial encoder with a bidirectional temporal module to map raw frames to
continuous intensity values.

\subsubsection{Spatial Encoder: ResNet18}

Each input frame $x_t$ is processed by a ResNet18 pretrained on ImageNet:
\begin{equation}
    \mathbf{f}_t = \mathrm{ResNet18}(x_t),
\end{equation}
producing a 512-dimensional feature vector.

\subsubsection{Temporal Encoder: Bidirectional GRU}

The sequence $(\mathbf{f}_1,\dots,\mathbf{f}_T)$ is passed to a bidirectional
GRU to capture forward and backward temporal dependencies:
\begin{equation}
    \mathbf{h}_t = \mathrm{BiGRU}(\mathbf{f}_1,\dots,\mathbf{f}_T).
\end{equation}
Experimental results indicate that the GRU captures most of the temporal
structure required for intensity estimation.

\subsubsection{Frame-Wise Regression Head}

A linear layer outputs the predicted intensity at each time step:
\begin{equation}
    \hat{y}(t) = \mathbf{w}^\top \mathbf{h}_t + b.
\end{equation}

\subsection{Training Objective}

The overall loss consists of a primary regression term and two optional temporal
regularizers.

\paragraph{Mean Squared Error (MSE)}
\begin{equation}
    \mathcal{L}_{\mathrm{mse}}
    = \frac{1}{T} \sum_{t=1}^{T} (\hat{y}(t) - y(t))^2.
\end{equation}

\paragraph{Smoothness Regularization}
\begin{equation}
    \mathcal{L}_{\mathrm{smooth}}
    = \frac{1}{T - 1} \sum_{t=1}^{T-1} (\hat{y}(t+1) - \hat{y}(t))^2.
\end{equation}

\paragraph{Apex Ranking Loss}
\begin{equation}
    \mathcal{L}_{\mathrm{rank}}
    = \max\left(
        0,\;
        1 - (\hat{y}(t_{\mathrm{ap}}) -
        \max_{t \ne t_{\mathrm{ap}}} \hat{y}(t))
    \right).
\end{equation}

\paragraph{Final Loss}
\begin{equation}
    \mathcal{L}
    = \lambda_{\mathrm{mse}} \mathcal{L}_{\mathrm{mse}}
    + \lambda_{\mathrm{smooth}} \mathcal{L}_{\mathrm{smooth}}
    + \lambda_{\mathrm{rank}} \mathcal{L}_{\mathrm{rank}}.
\end{equation}

Experiments suggest that the GRU alone captures most temporal structure, with
the auxiliary losses acting mainly as mild regularizers.

\subsection{Training Procedure}

Training is performed on a clip-wise basis using the Adam optimizer. Each batch
contains the temporally normalized frame sequence, its corresponding triangular
pseudo-label, and the predicted trajectory. Because all datasets follow the same
annotation schema and temporal normalization procedure, the training pipeline is
consistent and reproducible across datasets.

\section{Experiments}

This section presents the experimental setup used to evaluate the proposed 
framework for continuous micro-expression intensity estimation. All experiments 
use the unified preprocessing and temporal normalization procedure described in 
Section~\ref{sec:triangle}, ensuring consistent treatment across datasets. 
Performance is reported using Spearman’s $\rho$ and Kendall’s $\tau$, two 
correlation measures widely used to evaluate temporal agreement in intensity 
regression.

\subsection{Datasets}

We evaluate the method on two spontaneous micro-expression datasets that provide 
weak temporal annotations in the form of onset, apex, and offset indices.

\textbf{SAMM}~\cite{samm, davison2018objective, yap2020samm} contains 159 
micro-expressions recorded at 200\,fps with high spatial resolution. Each clip 
includes precise onset--apex--offset landmarks and annotated facial action units 
(AUs), making the dataset suitable for fine-grained temporal analysis.

\textbf{CASME~II}~\cite{casme2} consists of 247 spontaneous micro-expressions, 
also captured at 200\,fps, with detailed temporal annotations and AU labels. 
Following standard practice, only clips with unambiguous temporal landmarks are 
used.

Neither dataset provides frame-level intensity labels. Evaluation therefore 
relies entirely on triangular pseudo-intensity trajectories constructed from 
temporal landmarks, resulting in a fully weakly supervised and 
dataset-independent protocol.

\subsection{Evaluation Metrics}

For a predicted trajectory $\hat{y}(t)$ and its triangular pseudo-target 
$y(t)$, agreement is measured using:

\begin{itemize}
    \item \textbf{Spearman's rank correlation} ($\rho$), which assesses global 
    monotonic consistency across time.
    \item \textbf{Kendall's $\tau$}, which evaluates pairwise temporal ordering. 
\end{itemize}

Both metrics emphasize relative temporal structure rather than absolute 
intensity values, making them appropriate for weakly supervised regression.

\subsection{Training Details}

Models are trained with the Adam optimizer using the hyperparameters in 
Table~\ref{tab:hyperparams}. Each clip is uniformly resampled to $T$ frames 
between onset and offset, ensuring a consistent temporal length. Unless stated 
otherwise, we use $T{=}16$ frames of size $224{\times}224$.

\begin{table}[h!]
\centering
\caption{Training hyperparameters (fixed across all experiments).}
\label{tab:hyperparams}
\begin{tabular}{lc}
\hline
Frames per clip ($T$)       & 16 \\
Batch size                  & 8  \\
Learning rate               & $1\times10^{-4}$ \\
Backbone                    & ResNet18 \\
Temporal head               & Bidirectional GRU \\
Loss weights                & $\lambda_{\rm mse}{=}1$, $\lambda_{\rm smooth}{=}0.1$, $\lambda_{\rm_rank}{=}0.5$ \\
Optimizer                   & Adam \\
\hline
\end{tabular}
\end{table}

All model variants use identical resampling, augmentation, and optimization 
settings to ensure fair comparison.

\subsection{Ablation Studies}

We conduct ablations to assess the contribution of individual components:

\begin{enumerate}
    \item \textbf{Frame-wise Baseline (ResNet18)}: predicts each frame 
    independently without temporal modeling.

    \item \textbf{Full Temporal Model (ResNet18 + GRU)}: the proposed 
    spatial--temporal architecture.

    \item \textbf{Without Smoothness Loss}: removes $\mathcal{L}_{\mathrm{smooth}}$, 
    which encourages local temporal consistency. The GRU already enforces strong 
    smoothness, so the effect is minimal.

    \item \textbf{Without Apex Ranking Loss}: removes $\mathcal{L}_{\mathrm{rank}}$, 
    which encourages the apex to be the peak of the trajectory. CASME~II benefits 
    from removing this term due to weaker apex dominance.

    \item \textbf{Alternative Pseudo-Label Shapes (optional)}: evaluates Gaussian 
    and other shapes. The triangular prior consistently provides the most stable 
    supervision.
\end{enumerate}

Across all ablations, the GRU contributes the majority of performance gains, 
while auxiliary losses provide only modest refinements.

\subsection{Quantitative Results}
\label{sec:results}

We report results on SAMM and CASME~II using Spearman’s $\rho$ and Kendall’s 
$\tau$. All models are trained using the unified weakly supervised pipeline 
described in Section~\ref{sec:method}.

\subsubsection{Results on SAMM}

Table~\ref{tab:samm_results} summarizes the results. The frame-wise baseline 
achieves moderate agreement, indicating that spatial features alone are 
insufficient for capturing subtle micro-expression dynamics. Adding the GRU 
yields a substantial improvement of over \textbf{+0.16} in Spearman’s~$\rho$, 
with all temporal models exceeding 0.97.

Both auxiliary losses have limited impact. Removing $\mathcal{L}_{\mathrm{smooth}}$ 
slightly improves performance, while removing $\mathcal{L}_{\mathrm{rank}}$ has a 
negligible effect. These results suggest that the GRU provides sufficient 
inductive bias to learn smooth, peak-centered trajectories from the triangular 
prior.

\begin{table}[t]
\centering
\caption{Quantitative performance on the SAMM dataset.}
\label{tab:samm_results}
\begin{tabular}{lcc}
\hline
\textbf{Model} & \textbf{Spearman $\rho$} & \textbf{Kendall $\tau$} \\
\hline
ResNet18 (baseline)            & 0.8130 & 0.6697 \\
ResNet18 + GRU (full model)    & 0.9789 & 0.9222 \\
GRU w/o smoothness loss        & \textbf{0.9804} & \textbf{0.9252} \\
GRU w/o apex-ranking loss      & 0.9791 & 0.9221 \\
\hline
\end{tabular}
\end{table}

\subsubsection{Results on CASME~II}

Table~\ref{tab:casme2_results} presents the corresponding results on CASME~II. 
The GRU-based model again substantially outperforms the baseline, achieving 
correlations above 0.98 across ablations.

The best performance arises when $\mathcal{L}_{\mathrm{rank}}$ is removed. Many 
CASME~II clips exhibit weak or ambiguous apex frames, making a strict peak 
penalty unreliable. The GRU learns the dominant temporal transitions directly 
from the triangular prior, showing robustness to dataset-specific variability.

\begin{table}[t]
\centering
\caption{Quantitative performance on the CASME~II dataset.}
\label{tab:casme2_results}
\begin{tabular}{lcc}
\hline
\textbf{Model} & \textbf{Spearman $\rho$} & \textbf{Kendall $\tau$} \\
\hline
ResNet18 (baseline)            & 0.9172 & 0.7957 \\
ResNet18 + GRU (full model)    & 0.9884 & 0.9468 \\
GRU w/o smoothness loss        & 0.9876 & 0.9446 \\
GRU w/o apex-ranking loss      & \textbf{0.9894} & \textbf{0.9487} \\
\hline
\end{tabular}
\end{table}

\subsection{Qualitative Analysis}

\begin{figure}[H]
    \centering
    \includegraphics[width=0.85\linewidth]{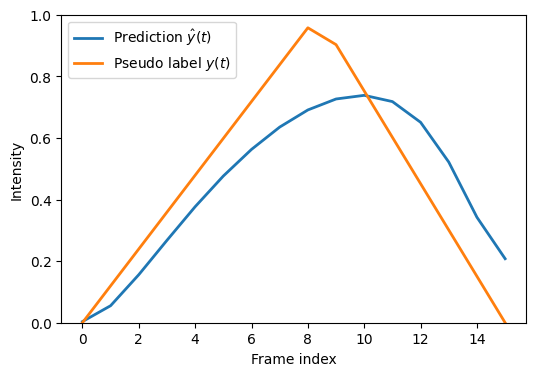}
    \caption{Qualitative examples comparing predicted intensity trajectories 
    (blue) with triangular pseudo-labels (gray). The GRU-based model produces 
    smooth and coherent temporal curves that align with expected 
    onset--apex--offset dynamics.}
    \label{fig:qualitative}
\end{figure}

Figure~\ref{fig:qualitative} shows qualitative examples of predicted trajectories 
alongside their triangular pseudo-labels. The GRU produces smooth, 
physiologically plausible curves that capture gradual onset, apex transitions, 
and natural decay, whereas the frame-wise baseline often produces noisy, 
incoherent predictions.

\section{Discussion}

The experimental results on SAMM and CASME~II indicate that the proposed
framework can estimate continuous micro-expression intensity under weak
supervision. Incorporating temporal modeling through a bidirectional GRU leads
to high agreement with the pseudo-intensity trajectories, with correlations
exceeding 0.97 on SAMM and 0.98 on CASME~II. These results suggest that
temporal modeling plays an important role in capturing the subtle dynamics of
micro-expressions when frame-level annotations are unavailable. The triangular
pseudo-label offers a simple and stable form of supervision that applies
consistently across datasets.

\subsection{Dataset-Specific Behaviour}

Although both datasets share the same onset–apex–offset annotation structure,
the ablation results show differences in how each responds to auxiliary losses:

\begin{itemize}
    \item \textbf{SAMM}: Removing either the smoothness or apex-ranking losses
    produces only small changes, with correlations remaining above 0.97. This
    suggests that the GRU alone captures most of the temporal structure in SAMM,
    whose clips typically exhibit clearer and more pronounced temporal patterns.
    Additional constraints appear to provide limited benefit once temporal
    modeling is in place.

    \item \textbf{CASME~II}: The highest performance is observed when the
    apex-ranking loss is removed. CASME~II clips tend to be shorter and may have
    less distinctive apex frames, making a strict peak-enforcing constraint less
    reliable. Allowing the GRU to learn the trajectory shape directly from the
    triangular prior yields more stable predictions. The smoothness term has
    only a minor effect as well.
\end{itemize}

Taken together, these observations indicate that the triangular pseudo-label
serves as a general supervisory signal that adapts well to dataset-specific
temporal characteristics without modifying the training procedure. The GRU
appears to account for most of the temporal consistency, while auxiliary losses
act primarily as mild regularizers.

\subsection{Cross-Dataset Generalization Considerations}

Although this study evaluates each dataset independently, the overall behaviour
of the model suggests that it would generalize well in cross-dataset settings.
The triangular pseudo-label normalizes temporal structure, and the temporal
model processes resampled sequences of fixed length, both of which help
mitigate differences in frame rates, clip durations, and subject identity. In
practice, SAMM and CASME~II differ in motion amplitude and apex sharpness, yet
the model achieves strong performance on both without dataset-specific
adjustments. This indicates that the proposed formulation may transfer well
across datasets with similar annotation formats, even though cross-dataset
experiments are beyond the scope of this study.

\subsection{Relation to AU Intensity Estimation Literature}

There is conceptual overlap between micro-expression intensity estimation and
AU-based intensity regression. Prior work in AU analysis often relies on dense
frame-level labels or structured priors to guide temporal learning. In contrast,
micro-expression datasets provide only sparse temporal cues. The present
framework aligns with AU regression methods in its use of temporal modeling,
but differs in its reliance on weak supervision and a simple triangular prior.
This distinction highlights the need for tailored approaches when dealing with
high-frame-rate micro-expressions, where dense annotations are rarely feasible.

\subsection{Limitations}

A key limitation of current micro-expression datasets is the absence of
frame-level intensity annotations. The proposed framework relies on triangular
pseudo-intensity trajectories derived from sparse temporal landmarks rather than
ground-truth intensity curves. While the pseudo labels provide useful structure,
they impose simplified assumptions about temporal progression and may not fully
capture cases in which real intensity profiles deviate from the idealized
triangular shape.

The lack of dense annotations also limits the ability to evaluate absolute
intensity accuracy. Future datasets incorporating human ratings, AU-based
intensity scores, or hybrid annotation strategies would allow for more direct
assessment. Despite these limitations, the results show that the framework
offers a practical and reproducible approach for continuous micro-expression
intensity estimation under realistic weak-supervision conditions.

\section{Conclusion}

This paper presented a framework for continuous micro-expression intensity
estimation under weak supervision using only onset–apex–offset annotations. The
approach combines a triangular pseudo-intensity trajectory with a
convolutional–recurrent model to produce dense frame-level predictions without
requiring manual intensity labels.

Experiments on SAMM and CASME~II show that the bidirectional GRU captures the
temporal structure characteristic of micro-expressions, achieving high agreement
with the pseudo-intensity trajectories. The triangular pseudo-label applies
consistently across datasets and remains effective even when auxiliary losses
are reduced or removed, suggesting that it provides a stable supervisory signal
in the absence of frame-level annotations.

Although the framework offers a practical and reproducible solution for
continuous intensity estimation, the lack of true frame-level intensity labels in
current micro-expression datasets remains a limitation. Additional annotation
resources—such as human intensity ratings, AU-based estimates, or other
complementary signals—would allow more direct evaluation and support further
refinement of temporal models. The formulation used here, however, is flexible
and could be extended to incorporate richer cues as they become available.

Overall, the method provides a dataset-independent approach for estimating
micro-expression intensity under realistic annotation constraints and offers a
basis for future work on fine-grained facial behaviour modeling, including
cross-dataset generalization and integration with broader affective computing
frameworks.

\bibliographystyle{ieeetr}
\bibliography{references.bib}

\end{document}